\documentclass[10pt,letter]{article}

\usepackage{cogsci}
\cogscifinalcopy
\usepackage{pslatex}
\usepackage{apacite}
\usepackage{graphicx}
\usepackage[font=small]{caption}
\usepackage{amsmath}

\usepackage{soul} 
\usepackage{color}
\usepackage{xcolor,colortbl}

\usepackage{subfigure}
\title{Learning word-referent mappings and concepts from raw inputs}
 
\author{{\large \bf Wai Keen Vong (waikeen.vong@nyu.edu) and Brenden M. Lake (brenden@nyu.edu}) \\
  Center for Data Science, 60 5th Ave \\
  New York, NY, 10011, USA} 

\begin{document}

\maketitle

\begin{abstract}

How do children learn correspondences between the language and the world from noisy, ambiguous, naturalistic input? One hypothesis is via cross-situational learning: tracking words and their possible referents across multiple situations allows learners to disambiguate correct word-referent mappings \cite{yu2007rapid}. However, previous models of cross-situational word learning operate on highly simplified representations, side-stepping two important aspects of the actual learning problem. First, how can word-referent mappings be learned from raw inputs such as images? Second, how can these learned mappings generalize to novel instances of a known word? In this paper, we present a neural network model trained from scratch via self-supervision that takes in raw images and words as inputs, and show that it can learn word-referent mappings from fully ambiguous scenes and utterances through cross-situational learning. In addition, the model generalizes to novel word instances, locates referents of words in a scene, and shows a preference for mutual exclusivity.

\end{abstract}

\section{Introduction}

Children must learn the meaning of words from noisy, sparse, and ambiguous information distributed across multiple modalities. Despite the computational challenges, children learn words at an impressive rate, estimated at upwards of ten per day on average between when they start speaking until the end of high school \cite{bloom2002children}. A key factor in understanding this efficiency is cross-situational learning: by tracking the co-occurrences between words and their referents across many individually ambiguous situations, learners can rapidly hone the meanings of words. Considerable evidence for cross-situational word learning has been found in laboratory studies of both adults \cite{yu2007rapid, kachergis2018word, yurovsky2015integrative} and in infants \cite{smith2008infants}.

There exist a variety of computational models of cross-situational word learning that provide theoretical accounts for the large body of empirical phenomena. Accounts based on ``associative learning'' track the observed statistics between words and referents across situations to determine the most plausible links \cite{kachergis2012associative, fazly2010probabilistic}. In contrast, accounts based on ``hypothesis testing'' consider only a limited number of hypotheses between words and possible referents \cite{trueswell2013propose}. A third set of models use Bayesian approaches to infer lexicons with high posterior probability, assuming that words are intentionally selected based on the objects in a given situation \cite{frank2009using, yurovsky2015integrative}.

Despite their successes, each of these models is limited by the simplicity of the assumed input representation. These models observe objects and words that are parsed from their raw forms and encoded into simplified symbolic representations that can be directly manipulated, side-stepping the question of how cross-situational learning can proceed from raw sensory inputs. Additionally, because these simplified representations are discretized, it becomes challenging to explain how learners can generalize to novel instances of words they have learned \cite{lewis2013integrated, taxitari2019limits}. A complete computational account of cross-situational word learning should explain generalization to novel instances of words, even when learning from the raw inputs of individually ambiguous scenes.

In this paper, we present a neural network model that learns word-referent mappings from ambiguous scenes presented as pixel-level images. We leverage recent ideas from self-supervised learning to train a model on a proxy supervised task and show that as a byproduct, the model learns representations that can detect the correct correspondences between objects and words. Our model is intended to be a computational-level account \cite{marr1982vision} of cross-situational word learning---demonstrating a means of solving the computational problems outlined above---rather than providing a step-by-step process-level account of learning. We show that our model can capture four kinds of phenomena related to word and concept learning: (1) learning correct word-referent mappings from fully ambiguous images and words; (2) generalizing to novel instances of words; (3) determining the particular location of a referent in a scene; and (4) generalizing with a preference for mutual exclusivity.

\section{Model}

\begin{figure}[th]
\begin{center}
\includegraphics[width=0.85\linewidth]{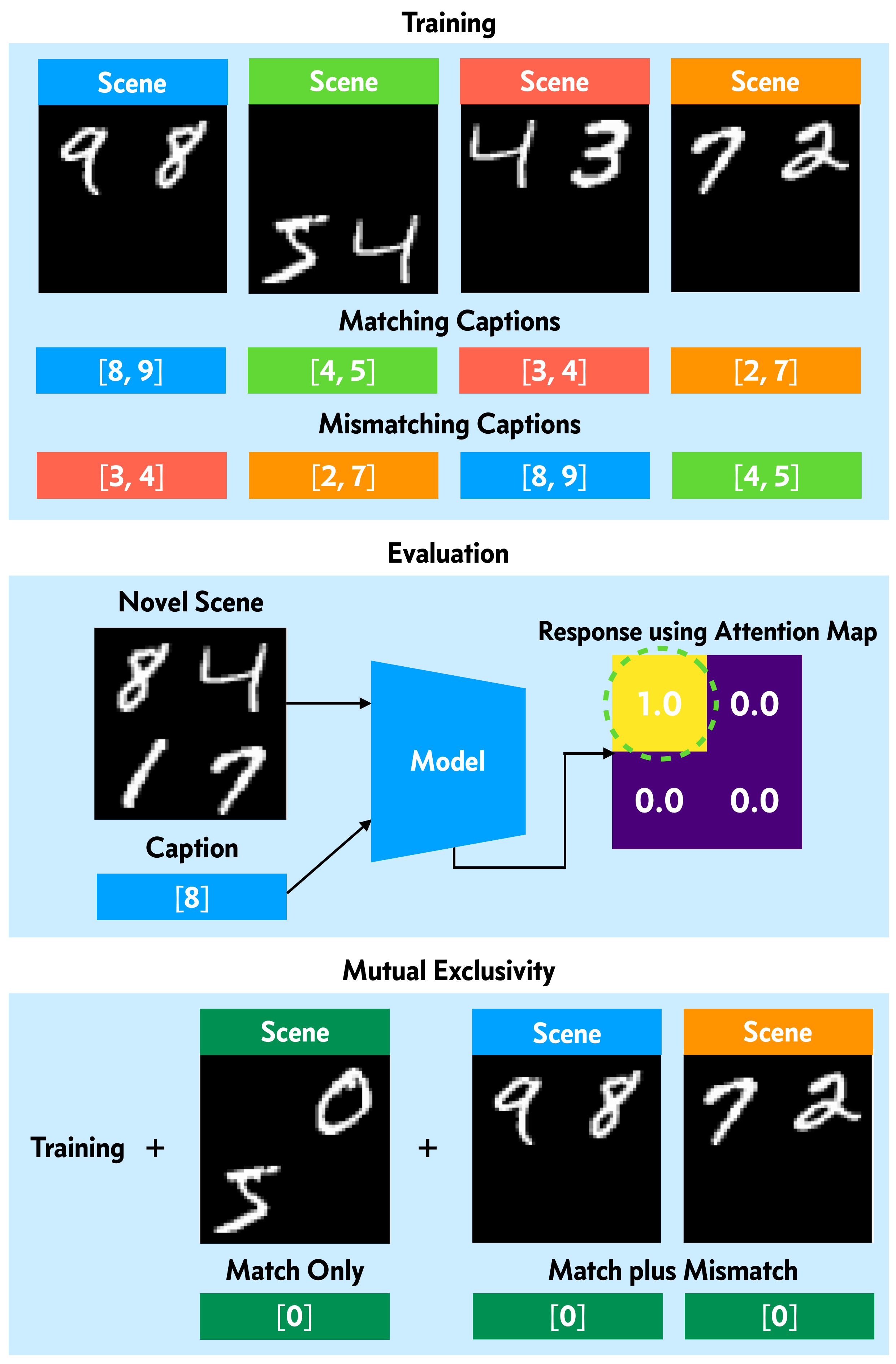}
\end{center}
\vspace{-0.25cm}
\caption{\textbf{Scenes and captions used for training (top), evaluation (middle) and testing for mutual exclusivity (bottom).} During training, the model is presented with a set of matching scenes and captions (illustrated with the same color), containing MNIST digits arranged in a 2x2 grid. Mismatching captions are created by permutation, and the model learns to discriminate between matching vs. mismatching scenes and captions. During evaluation, the trained model sees a novel scene and a single target word, and selects the location of the attention map with the highest value as its response for determining the target referent. To test for mutual exclusivity, we created a separate training set of scenes and captions that excluded a single digit, and then added either a single matching scene and caption containing the novel digit (0), or provided five additional mismatches, and examined the model's preference for the novel digit after training.}
\label{matching-vs-mismatching-scenes}
\vspace{-0.25cm}
\end{figure}

Our model takes two inputs: an image of a \textbf{scene} containing some number of objects, and a \textbf{caption} containing an array with the same number of words. Example scenes and captions are illustrated in the top of Figure~\ref{matching-vs-mismatching-scenes}. The caption associated with each scene can be either \textit{matching}, where all of the words match with the objects in the scene, or \textit{mismatching}, where at least one of the words does not match with one of the objects in the scene.

During learning, the model is trained to predict whether or not a scene and caption match. Our aim is that training on this discrimination task will produce representations that properly disambiguate the correct word-referent mappings from the training data. However, this set-up raises an important question: where do mismatching scenes and captions come from, if a child only experiences positive examples (matching scenes and captions)? One possibility is that the learner can implicitly construct mismatching scenes by replaying past scenes and combining them with captions from other scenes, as illustrated in Figure~\ref{matching-vs-mismatching-scenes}.\footnote{This method of generating mismatching scenes assumes that the only words that match a given scene are the ones it was originally paired with, and not any other caption when performing this mismatching procedure, and provides a slight inductive bias towards mutual exclusivity.} As the process of generating mismatching trials involves recombining the input data in a novel manner, and turning this from an unsupervised learning task to a supervised learning task, this is a form of \textit{self-supervised learning} \cite{goyal2019scaling}.

\begin{figure}[t]
\begin{center}
\includegraphics[width=\linewidth]{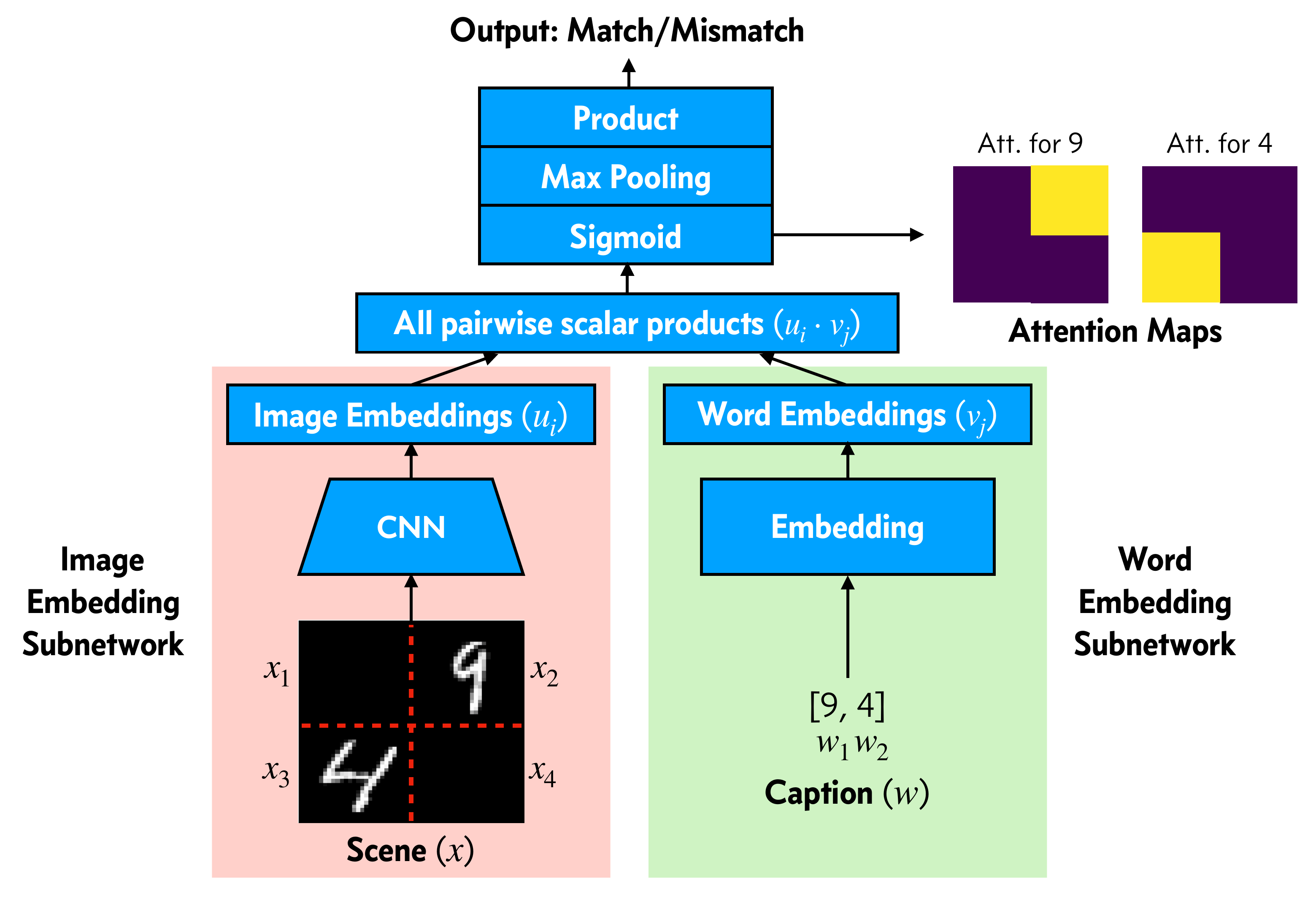}
\end{center}
\vspace{-0.25cm}
\caption{\textbf{Model Architecture}. The network takes a scene (splitting it into four quadrants) and a caption as inputs, and embeds the information from each of these modalities using a CNN and Embedding layer respectively. It then applies a pairwise scalar product operation to detect correspondences, combining this information to produce attention maps that can be used to visualize word-referent mappings. As output, the network predicts whether a given scene and caption are matching or mismatching.}
\vspace{-0.25cm}
\label{model-architecture}
\end{figure}

The model architecture is inspired by the Audio-Visual Object Localization model (AVOL-net) from \citeA{arandjelovic2018objects}. In their work, the architecture is used to train a model that detects correspondences between images and audio sounds (not necessarily language), and can determine which parts of a particular image the sound may have come from. Using this as a starting point, we modified the architecture to match the designs used to study cross-situational word learning. A figure depicting the key elements of the architecture are shown in Figure~\ref{model-architecture}.

\subsubsection{Image and word embeddings.} The scene and caption are first processed separately using an \textit{image embedding subnetwork} (Figure~\ref{model-architecture}; red shading) and \textit{word embedding subnetwork} (green shading), respectively. As a pre-processing step, each scene $x$ is broken up into four quadrants of equal size, and each 28x28px image $x_i$ is passed into the image embedding subnetwork, although segmentation is not compulsory in this framework.\footnote{Segmentation was found to reduce the sample complexity of cross-situational learning. One of our analyses compares this model to one without this pre-processing step.} This subnetwork is a convolutional neural network, that outputs four separate image embedding vectors $u_i \in \mathbf{R}^d$, where $d$ is the dimensionality of the embedding space. Because the same convolutional neural network is used to process each part of the image separately that contains the various objects, we call this model the ``Object-CNN Model''. Concurrently, each of the words $w_j$ in the caption $w$ is passed into the word embedding subnetwork (consisting of a single Embedding layer denoted as $f_e$) such that each word is represented by a word embedding vector $v_j \in \mathbf{R}^d$. The dimensionality of the word embedding vectors are designed to have the same dimensionality as the image embedding vectors. These operations are notated as follows,
\begin{equation}
    u_i = \text{CNN}(x_i) \text{ and } v_j = f_e(w_j).
\end{equation}

\subsubsection{Attention maps.} After computing the embeddings for each modality, the model computes a correspondence score $s_{ij} \in \mathbf{R}$ between each image embedding $u_i$ and word embedding $v_j$ via a scalar product operation. We divide the correspondence score by the square root of the size of the embedding dimensionality \cite{vaswani2017attention} and then apply a sigmoid operation to produce a bounded scalar attention score $a_{ij} \in [0, 1]$,
\begin{equation}
    s_{ij} = u_i \cdot v_j \quad \text{ and } \quad a_{ij} = \sigma(\frac{s_{ij}}{\sqrt{d}}).
\end{equation}
Using these attention scores, we can produce an \textbf{attention map} by concatenating all of the attention scores for a given word $w_j$, that depicts where the model believes the referents for each word are located in the scene (Figure~\ref{model-architecture}; heatmaps for each label).

\subsubsection{Output.} Finally, by applying a max operator over the attention map for each word results in the sub-output $o_j \in [0, 1]$, which represents the probability that the word $w_j$ was detected in the scene. The final output of the model $o \in [0, 1]$ is simply a product of all of the sub-outputs, reflecting the fact that for a match response, every word needs to be matched with a corresponding object in the scene,
\begin{equation}
    o_j = \max_i a_{ij} \quad \text{ and } \quad o = \prod\limits_{j=1}^k o_j.
\end{equation}
The model receives the correct response (match or mismatch) as binary feedback. More importantly, it does not receive any additional feedback for how the attention maps between words and objects should be organized, and must learn to find good representations to achieve this. Our goal is that by training the model on this discrimination task, we can investigate whether the learned representations can isolate the referents for each word in a manner that demonstrates cross-situational learning.

\section{Simulations}
We report extensive cross-situational learning simulations that vary key difficulty factors including scene complexity, generalization to new exemplars, and amount of training. All of the simulations were based on a synthetic dataset with generated scenes and captions, providing us with experimental control over all aspects of the evaluation. Scenes were 56x56px in size, and contained objects in some of the quadrants of the scene. The objects used in the scenes were digits from MNIST \cite{lecun1998gradient}, a database of thousands of handwritten digits and a commonly used dataset in machine learning. For each scene, we also generated captions that were \textbf{matching}, where the words in the caption matched the digits that appeared in the scene. We then generated an equivalent number of \textbf{mismatching} scenes and captions by switching the captions from its paired matching scene to a different scene, as illustrated in Figure~\ref{matching-vs-mismatching-scenes}.\footnote{This method of generating mismatching scenes assumes that the only words that match a given scene are the ones it was originally paired with, and not any other caption when performing this mismatching procedure.} We tested the model along three conditions:
\begin{itemize}
    \item \emph{Scene complexity}: The referential ambiguity on each trial was manipulated, ranging from 2 digits per scene and 2 words per caption (\textsc{two-objects}), 3 digits and 3 words (\textsc{three-objects}), or 4 digits and 4 words per trial (\textsc{four-objects}). 
    \item \emph{Generalization type}: In the \textsc{fixed} condition, the same fixed instance of each digit was used in training and evaluation, requiring the network to generalize only to novel scenes that combine known digit instances in new ways.\footnote{Arbitrarily chosen as the first instance of each digit in the MNIST training set.} On the other hand, in the \textsc{varying} condition, the model was presented with varying instances of the same digit sampled from the training set of MNIST, requiring the model to handle both new scenes and new instances. During evaluation, digit instances were chosen from the MNIST test set to ensure novelty.
    \item \emph{Training set size}: The amount of matching word-object pairs presented to the network was also varied. In the \textsc{fixed} example types, the model was presented with 36 to 720 matching word-object pairs, although in some cases we were limited by the number of possible unique combinations of captions that could be generated for some of the difficulty conditions. In the \textsc{varying} simulations, the model was presented with 36 to 3600 matching word-object pairs. An equal number of mismatching word-object pairs were also generated using the procedure described above, although in principle, many more mismatching scene and caption pairs could potentially be generated. Additionally, because we controlled for the number of matching word-object pairs, the exact number of scenes and captions for the same training set size varied across scene complexity conditions.
\end{itemize}

\subsubsection{Training details.} A few additional details are required to describe how we trained the model. The embedding dimensionality for both the image and word embedding subnetworks was set to be 64, the input size of the word embedding subnetwork was 10 (matching the number of possible digits) and the weights initialized as an identity matrix. The convolutional neural network consisted of two convolutional layers (with 16 and 32 feature maps) with 2x2 max pooling layers, followed by two fully connected layers. All of the activations in the convolutional layers and the first fully connected layer were ReLU activations, with a dropout layer (set to 50\% dropout) in between the two fully connected layers. The model was trained for 1000 epochs, with 5 independent runs for each condition and the results averaged. The learning rate was set to 3e-4 and trained end-to-end with the AdamW optimizer \cite{loshchilov2017decoupled},  using a binary cross-entropy loss along with weight decay of 1e-4. The batch size used for training was 12 for all simulations, except for the varying example conditions with more than 360 matching word-object pairs, where the batch size was increased to 120. The model is trained from scratch, rather than using existing pre-trained representations that were trained in a supervised setting, to demonstrate that our model can indeed perform cross-situational word learning from ambiguous scenes and captions only. Additionally, the model is trained in batch, rather than on-line, as we are aiming at a computational-level rather than process-level account. Thus the network does not automatically make trial-by-trial predictions regarding behavior, similar to the model presented in \citeA{frank2009using}, but instead is compared on evaluation performance after training. Alternatively, behavior for varying amounts of experience can be modeled as varying the training set size, as described above.

\begin{figure}[t!]
\begin{center}
\includegraphics[width=\linewidth]{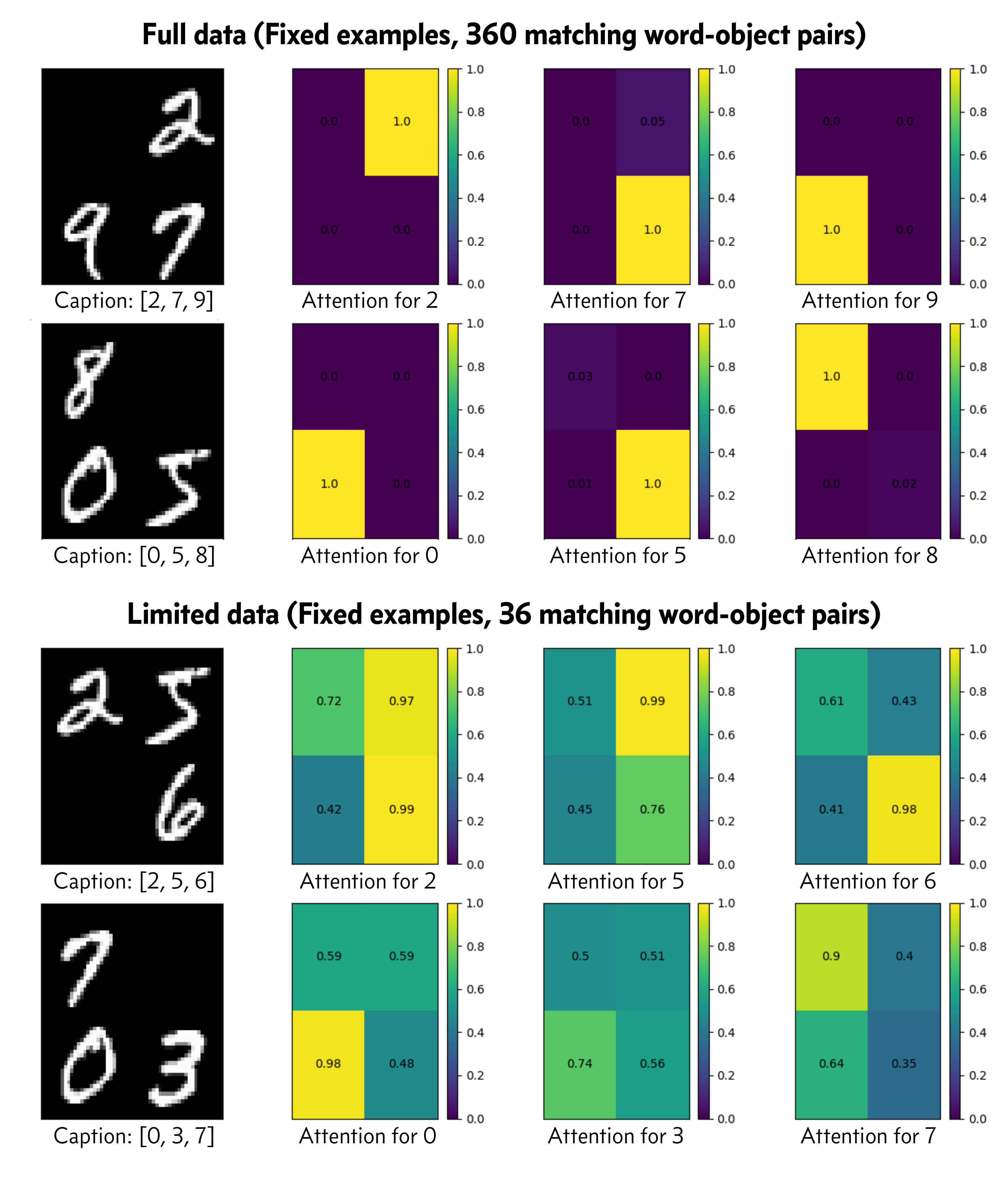}
\end{center}
\vspace{-0.25cm}
\caption{\textbf{Examples of attention maps produced by the model}. Each row shows the scene (left), along with the three associated attention maps for each word in the caption for each scene (right). Lower attention scores are in purple, while higher scores are in yellow. The first two rows show that sufficient data produces highly peaked attention maps for each word in the correct locations of the scene, while the bottom two rows show that limited data results in some incorrect correspondences and fuzzier attention scores.}
\label{attention-maps}
\vspace{-0.5cm}
\end{figure}

\section{Results}

To begin, we examine whether our model learns the proxy discrimination task it is trained on. The model's training accuracy in the final epoch of training was 97.5\%, when averaged across the different conditions, suggesting that it effectively learns to discriminate between matching and mismatching trials observed during training. However, and more importantly, does this result in representations that demonstrate the model isolates the correct word-referent mappings? We address this with two analyses: a qualitative analysis by presenting some attention maps generated by the model, and a more thorough quantitative analysis to evaluate the performance of the model across the various simulation conditions using these attention maps. We also present results from two follow-up simulations investigating whether our model displays mutual exclusivity, and compare our model to a variant that does not require the pre-processing step that segments scenes.

\subsubsection{Attention maps.} Although the model discriminates between matching and mismatching training trials, does it do so by learning representations that isolate the correct underlying word-referent mappings, or by some other strategy? To investigate this, we can start by looking at the intermediate computations that produce the attention maps for each word. As shown in Figure~\ref{attention-maps}, the attention map for each word visualizes the degree of correspondence the model thinks exists between each word embedding and each image embedding from the four quadrants of each scene. This provides us with a qualitative sense of the model's behavior, and where it looks at a given scene when presented with each word. 

From Figure~\ref{attention-maps}, we see that when the model is presented with a sufficient number of scenes and captions for training, the model learns the correct word-referent mappings where the attention is only active for the part of the scene containing the word, and zero otherwise. However, with a limited number of scenes and captions presented to the model during training, the model is both not confident about particular correspondences (even though they may be in the correct direction), and also fails to rule out incorrect associations between some words and referents. \\
    
\subsubsection{Mapping evaluation.} We also performed a more rigorous quantitative evaluation of the model's ability to learn word-referent mappings. Mimicking the evaluation procedure for behavioral experiments with children and adults \cite{yu2007rapid, smith2008infants}, we generated novel evaluation scenes that consisted of a target digit, along with three foil digits chosen at random, and paired it with a caption containing a single word corresponding to the target digit, as illustrated in the bottom part of Figure~\ref{matching-vs-mismatching-scenes}. This four alternative forced-choice procedure was used regardless of the scene complexity seen at training. Each evaluation scene and caption was passed into a trained model, and used the location of the maximum value in the attention map produced to determine the model's response. If the position of the maximum attention value was the same as the location of the target digit, this indicates that the model had indeed learned the correct word-referent mapping.

\begin{figure}[t!]
\begin{center}
\includegraphics[width=\linewidth]{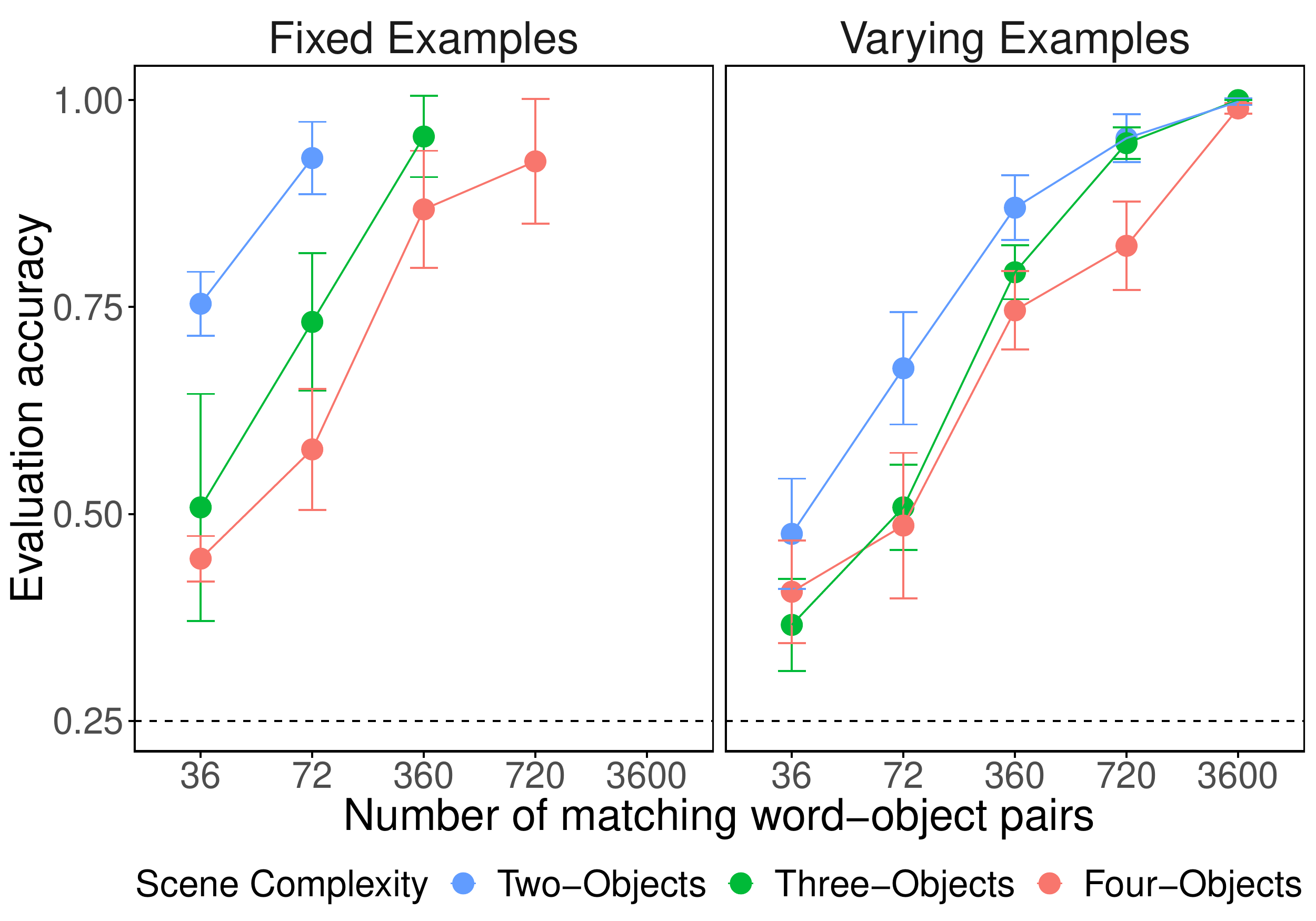}
\end{center}
\vspace{-0.25cm}
\caption{\textbf{Evaluation performance of the Object-CNN model}. Accuracy as a function of the number of matching word-object pairs, with color representing the difficulty condition (number of objects in each scene). Error bars are 95\% confidence intervals, and the dashed line indicates chance performance.}
\label{multi-evaluation-performance}
\vspace{-0.5cm}
\end{figure}

We performed 100 evaluations for each trained model (consisting of ten separate evaluations per target word with different foils) across all conditions. Results are shown in Figure~\ref{multi-evaluation-performance}, where each plot shows the evaluation accuracy (averaged across the five training runs of each condition with different random seeds). Overall, we see that performance improves as more word-object pairs are observed for both the fixed and varying example types, with evaluation accuracy scores greater than 90\% in each condition. Furthermore, the high evaluation accuracy achieved in the varying condition highlights our model's ability to generalize, as it suggests that for a learned word, the model can determine the correct referent using novel examples of that word that were not seen during training. However, learning to generalize concepts to novel instances requires an order of magnitude more data (3600 matching word-object pairs) than merely learning the correct mappings in novel scenes.

We also find that evaluation performance consistently decreases with increasing scene complexity (increasing ambiguity per scene), matching empirical studies of cross-situational word learning with adults \cite{yu2007rapid}. Furthermore, although the experimental design of \citeA{yu2007rapid} was slightly different to our simulations (18 different categories with 54 word-object pairs), they observed evaluation accuracy scores of 89\%, 72\% and 56\% for the Two-Object, Three-Object and Four-Object conditions respectively in their task. (fixed examples only). We observed a qualitatively similar pattern of results in the limited data case with fixed examples, with evaluation performance of 93\%, 72\% and 58\% respectively with 72 matching word-object pairs. It is surprising that the neural network and human participants achieve similar levels of accuracy for a limited number of word-object pairs, given that neural networks are notoriously data hungry \cite{geman1992neural}.

\subsubsection{Mutual exclusivity.}

One of the hallmarks from both children's early word learning and empirical findings of cross-situational word learning is mutual exclusivity, the preference to map a novel word onto a novel object \cite{markman1988children, lewis2020role}. Many other models of cross-situational word learning can account for mutual exclusivity \cite{kachergis2012associative, frank2009using, mcmurray2012word} although standard deep neural networks struggle with this type of reasoning \cite{gandhi2019mutual}, and thus we were interested in examining whether our model captures this important phenomenon. We conducted a separate set of simulations to determine whether or not our model displays a preference for mutual exclusivity. 

We generated a separate set of training trials consisting of 72 matching word-object pairs, with two objects per scene using fixed examples, and excluded one digit from both scenes and captions from this training set.\footnote{This condition was chosen since this amount of training data was sufficient for very high evaluation performance in our main simulations.} To evaluate mutual exclusivity, the model was also provided with an additional training trial (or trials) involving the novel digit paired with a familiar foil digit, simulating the developmental paradigm where children are asked to ``Show me the dax'' when given a novel and familiar object \cite{markman1988children} (see Figure \ref{matching-vs-mismatching-scenes} bottom). In the \textit{Match Only} condition, we provided the model with this single additional mutual exclusivity trial, consisting of a novel digit and a foil digit along with the novel word as the caption, treating this as a matching trial. In the \textit{Match plus Mismatch} condition, in addition to the single matching trial with the novel digit and word, we also paired the caption containing the novel word with five of the other training scenes (that only contained other digits) to create additional mismatched trials. The models were trained for 500 epochs, rather than 1000 epochs, but was otherwise trained in exactly the same manner as described earlier. For each excluded digit, we performed 10 independent runs.

To determine whether or not a trained model displays a preference for mutual exclusivity, we first examined whether the model produced the correct match output response for the mutual exclusivity trial, and from this, calculated the proportion of simulations where the model's attention for the novel word was higher for the novel digit than for the foil digit. In the \textit{Match Only} simulations, the model's preference for the novel digit was 51\%, suggesting that providing this single additional matching trial did not result in any preference for mutual exclusivity. On the other hand, the \textit{Match plus Mismatch} simulations showed a greater preference for mutual exclusivity, with 73\% of runs that favored the novel digit. This suggests that augmenting the model with a few additional mismatched trials with the novel word was sufficient to induce mutual exclusivity in our model, without requiring any additional changes.\footnote{In the simulations for the \textit{Match plus Mismatch} condition, we observed some instabilities during training that led to around 5\% of runs showing a preference for the blank quadrants, rather than the novel or foil digit, which were excluded when calculating the preference for ME.}

\subsubsection{Cross-situational learning without segmentation pre-processing.}

\begin{figure}[t!]
  \centering
  \includegraphics[width=.9\linewidth]{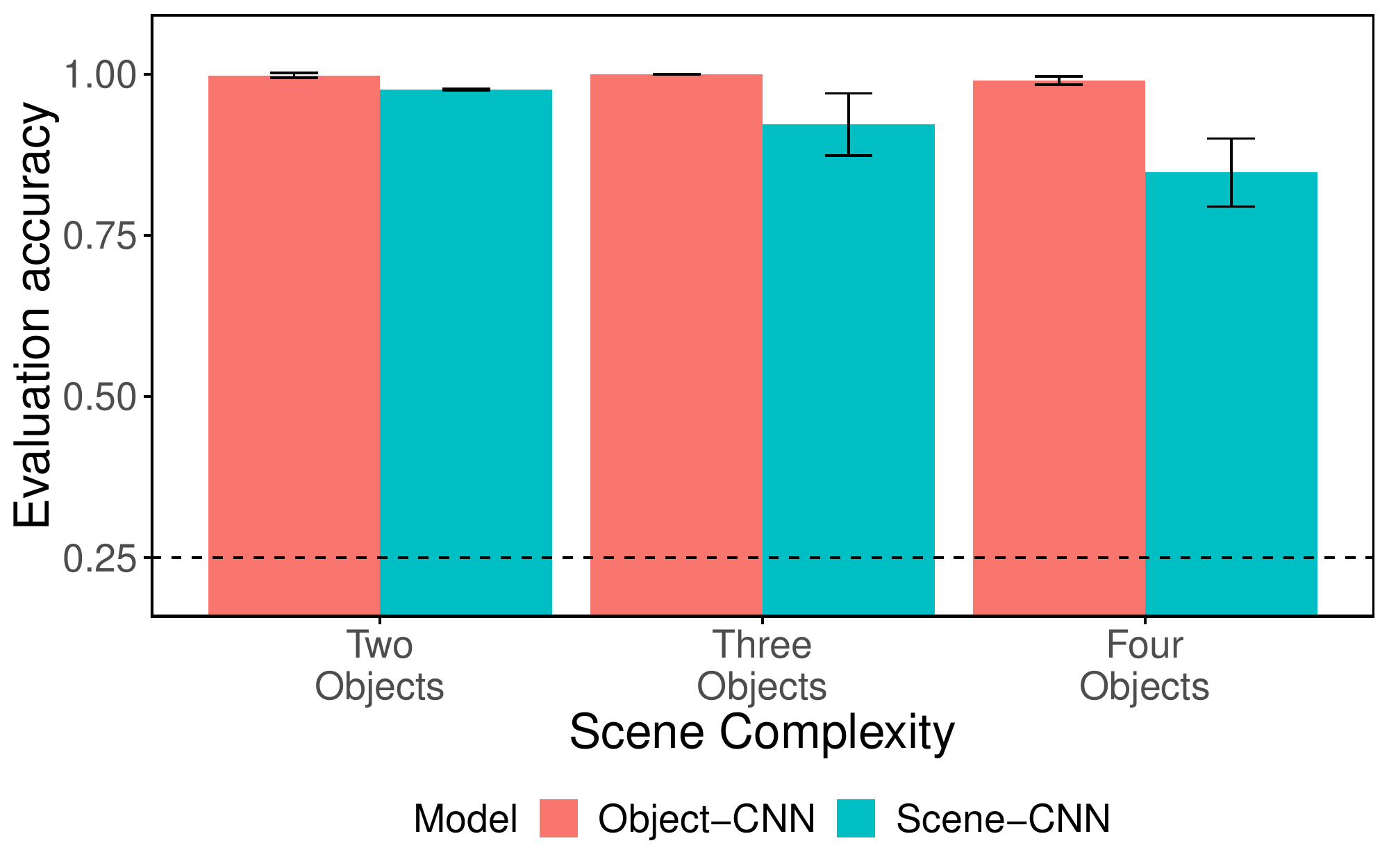}
\vspace{-0.25cm}
\caption{\textbf{Evaluation accuracy comparing the Object-CNN and Scene-CNN architectures}, with error bars showing 95\% confidence intervals. Results shown are from the 3600 matching word-object pairs with varying examples (dashed line indicates chance performance), showing that while the Scene-CNN model can identify word-referent pairs after training without pre-processing the objects in the scene, its performance decreases with increasing ambiguity.}
\label{object-vs-scene-cnn}
\vspace{-0.5cm}
\end{figure}

A stated goal of this work is to learn word-referent mappings from raw images, but the networks so far rely upon pre-processing to segment raw scenes into a set of candidate referent images. Children are not provided with such signals in realistic learning environments, and for the final analysis, we examine whether or not this pre-processing is essential to our approach. Instead of the Object-CNN model, we consider a Scene-CNN model that takes in the full 56x56px scene as a single input, and applies a different convolutional neural network that outputs four image embeddings $u_i$ as before. The Scene-CNN model was trained in exactly the same manner as the Object-CNN model described earlier on the 3600 matching word-object pair condition with varying examples, and then performed the same evaluation as described earlier. The results are shown in Figure~\ref{object-vs-scene-cnn}, and show that while the performance of the Scene-CNN model decreases more than the Object-CNN model with the increasing scene complexity, performance is still far above chance suggesting that it can also learn correct word-referent mappings without additional segmentation.\footnote{The Scene-CNN model was also tested with the smaller training set sizes but evaluation performance was much lower than the Object-CNN model, despite achieving similar discrimination performance during training. This suggests that the inductive bias from extracting the objects in the Object-CNN model greatly helps in learning representations that lead to cross-situational word learning.} Crucially, it shows that the approach can perform cross-situational learning from raw, unsegmented images of the scene. An interesting open question is whether an architecture that first performs object detection, or one that operates over the entire scene would scale to more realistic kinds of naturalistic data a child may encounter. 

\section{Discussion}

We present a computational-level account of cross-situational word learning from images and words using a self-supervised learning approach. Our model provides a computational account for learning word-referent mappings from ambiguous raw inputs (images and words), and shows generalization to novel scenes and novel exemplars of these learned words, feats not been achieved by other models of cross-situational word learning. In addition, we can localize the intended referent from a given scene through the attention maps, and show that the model displays a slight preference for mutual exclusivity. 

While our work provides a proof-of-concept that cross-situational word learning can be achieved from raw inputs, there are a number of limitations of the current model due to the idealized set-up of our training procedure, but might be interesting directions for future research. First, the model requires the number of objects in a scene to be the same as the number of words, and relaxing this assumption may require other methods of detecting correspondences across modalities that are more graded than a binary match or mismatch response. Second, in our data generation process, we did not consider the effect of noise, where matching scenes and captions may have errors \cite{fazly2010probabilistic}. Finally, our model currently takes in words as text while cross-situational word learning experiments often provide participants the words as audio. One possible method for capturing this additional detail would be to replace the word embedding layer with a second convolutional neural network that takes in an audio spectrogram as input, as demonstrated by \citeA{arandjelovic2018objects}. 

This work demonstrates how simultaneous cross-situational word learning and concept learning is possible with raw inputs from scratch, yet more work is needed before models of word learning in the lab generalize to word learning in the wild. Unlike our model, by the time children start learning words they also have access to object representations and the ability to segment objects and words, and these richer representations and abilities may be advantageous for word learning \cite{lake2017building}. One avenue for future work is to try and scale up our model to naturalistic, longitudinal video headcam datasets \cite{sullivan2020saycam}. These datasets provide rich and detailed access to the kinds of environmental statistics that children receive from a first-person perspective, and such datasets may help determine the necessary computational machinery required for word learning at scale.



\bibliographystyle{apacite}

\setlength{\bibleftmargin}{.125in}
\setlength{\bibindent}{-\bibleftmargin}

\bibliography{references}

\end{document}